\documentclass[11pt,a4paper]{article}
\usepackage{authblk}
\usepackage[hyperref]{emnlp2020}
\usepackage{times}
\usepackage{latexsym}

\usepackage{microtype}

\usepackage{makecell, multirow, graphicx}
\usepackage{amsmath}
\usepackage{amssymb}
\usepackage{xcolor}

\DeclareMathOperator*{\argmax}{argmax}

\aclfinalcopy 

\title{Interpreting Verbal Metaphors by Paraphrasing}

\author[1]{\textbf{Rui Mao}}
\author[2]{\textbf{Chenghua Lin}}
\author[3]{\textbf{Frank Guerin}}
\affil[1]{Department of Computing Science, University of Aberdeen, AB24 3UE, UK} 
\affil[1]{\vspace*{0.1cm}\tt {r03rm16@abdn.ac.uk}}
\affil[2]{Department of Computer Science, University of Sheffield, S1 4DP, UK}
\affil[2]{\tt {c.lin@sheffield.ac.uk}}
\affil[3]{Department of Computer Science, University of Surrey, GU2 7XH, UK}
\affil[3]{\vspace*{0.1cm}\tt {f.guerin@surrey.ac.uk}}

\date{}

\begin{document}
\maketitle
\begin{abstract}
  Metaphorical expressions are difficult linguistic phenomena, challenging diverse Natural Language Processing tasks. 
  Previous works showed that paraphrasing a metaphor as its literal counterpart can help machines better process metaphors on downstream tasks. 
  In this paper, we interpret metaphors with BERT and WordNet hypernyms and synonyms in an unsupervised manner, showing that our method significantly outperforms the state-of-the-art baseline. We also demonstrate that our method can help a machine translation system improve its accuracy in translating English metaphors to 8 target languages.
\end{abstract}

\section{Introduction}

Metaphor is defined as using one or several words to illustrate a  meaning different to the basic meaning of the words \citep{steen2010method}. Due to the difficulty of inferring underlying meanings of metaphors, metaphoric expressions challenge diverse Natural Language Processing (NLP) tasks, e.g., sentiment analysis \citep{ghosh2015semeval} and machine translation (MT) \citep{mao2018word}.
Currently, BERT has achieved large improvements on diverse downstream tasks \citep{devlin2019bert}. 
However, it has not been examined on metaphor interpretation.
In this paper, we interpret verbal metaphors with BERT and WordNet \citep{fellbaum2005wordnet}. We focus on verbal metaphors, because verbs are the most common metaphoric expressions among all PoS categories \citep{steen2010method},
forming the most popular tasks in automatic metaphor identification and interpretation \citep{shutova2015design,leong2018report}.

We consider metaphor interpretation as a paraphrasing task, 
predicting literal counterparts for metaphors. Unlike fine-tuning or feature based BERT applications, we use BERT as a language modelling (LM) method to predict a semantically similar word to the real meaning of a metaphor, which has the highest probability of appearing in a given context as the literal paraphrase of the metaphor (see \S~\ref{sect: Methodology} for the reason). 
The probability is given by BERT based LM prediction and the semantically similar word is constrained by WordNet hypernyms and synonyms. Therefore, our approach is fully unsupervised which does not require labelling a large dataset for model training.

By running an automatic evaluation on a publicly available dataset \citep[MOH,][]{mohammad2016metaphor}, our model achieves 8\% accuracy gains on verbal metaphor paraphrasing, compared with the state-of-the-art baseline \citep{mao2018word}. On the human evaluated paraphrasing task, the average gain of our model is 13\% on two benchmark datasets, i.e., MOH and VUA \cite{steen2010method}. We also examine the metaphor paraphrases on MT tasks based on Google translator\footnote{\url{https://translate.google.com/}} and 8 diverse target languages. Evaluation results show that our method can improve the accuracy of translating English metaphors by an average gain of 20.9\%, outperforming the baseline by 12.9\%.

The contribution of this work can be summarised as follows: (1) we introduce a simple yet effective method by using BERT as a LM model for metaphor interpretation, yielding significant improvements against the state-of-the-art baseline; (2) with our model, a MT system can significantly improve its accuracy of translating English metaphors into 8 diverse target languages.

\section{Related work}

Due to the lack of large annotated metaphor interpretation corpora, previous works addressed metaphor interpretation mainly by unsupervised modelling of co-occurrences of words and their contexts with different knowledge bases and lexical resources \citep{shutova2010automatic,shutova2012unsupervised,bollegala2013metaphor}. These works focused on interpreting verbal metaphors from word-pairs with specific syntactic structures, e.g., verb-subject and verb-direct object, which is inconvenient for real-world applications. Another trend in metaphor interpretation focused on a specific domain, e.g., question answering about Unix \citep{martin1990computational}, event descriptions in economics \citep{narayanan1997knowledge} and mental states descriptions \citep{barnden2002artificial} with hand-coded knowledge and logic rules. \citet{mao2018word} extended the word-pair models, interpreting metaphors from full sentences in open domains. They modelled the co-occurrences of words and their contexts with word2vec CBOW \citep{mikolov2013distributed} input and output vectors to generate appropriate paraphrases of metaphors. 
Compared with context-independent word2vec embeddings, BERT as a context-dependent word embedding method, has shown remarkable performance on diverse NLP tasks \citep{devlin2019bert}. However, to the best of our knowledge, BERT has not been applied in metaphor interpretation. 

\section{Methodology}
\label{sect: Methodology}
Inspired by \citet{mao2018word}, we introduce a hypernym and synonym constrained missing word prediction method for unsupervised metaphor interpretation. The difference is our missing word prediction is based on pre-trained BERT (\textsc{bert-large-cased}\footnote{\url{https://github.com/google-research/bert}}) that has 24 Transformer layers \citep{vaswani2017attention}, whereas \citet{mao2018word} used word2vec input and output vectors.

BERT is a pre-trained Language Modelling method. The training target is to predict randomly selected masked WordPieces \citep{wu2016google} and the next sentence of a current processing sentence. Since the masked word prediction is bidirectional, combining its surrounding context information, a pre-trained BERT model on open corpora can be naturally applied as a missing word prediction model. In our task, we predict the probability of a missing word ($m_{i}$) with its context ($w$) by using pre-trained BERT
\begin{align*}
\begin{gathered}
    p(m_{i}|w_{1}, ..., w_{i-1}, w_{i+1}, ..., w_{n})\\
    =  \text{BERT}(\textsc{[cls]}, w_{1}, ..., \textsc{[mask]}_{i},..., w_{n}, \textsc{[sep]}),
\end{gathered}
\end{align*}
where the position of $m_{i}$ is replaced with $\textsc{[mask]}_{i}$; \textsc{[cls]} and \textsc{[sep]} are special tokens, representing the start and end of an input sequence with a length of $n$. For a sentence that has multiple metaphors, we mask one metaphoric word each time.

Next, we predict an appropriate paraphrase for the metaphor. In order to connect a metaphor (the target word) with its possible literal counterparts, we introduce the semantic constraints of WordNet hypernyms and synonyms for candidate literal counterpart development. It is likely that one of the hypernyms and synonyms has a similar meaning to the underlying meaning of a metaphor, thus, they help the BERT model filter out irrelevant predictions. The hypernyms $\{h_{i,j}\}$ and synonyms $\{s_{i,k}\}$ of a metaphoric verb $i$ and their different verb forms ($f$, inflections) are considered as our candidate word set ($\mathcal{C}_{i}=\{h_{i,j}, h^{f}_{i,j}, s_{i,k}, s^{f}_{i,k}\}$). 

The best fit word ($b_{i}$) in $\mathcal{C}_{i}$ is given by a candidate word with the highest probability in its context
\begin{equation*}
    b_{i} = \argmax_{m_{i} \in \mathcal{C}_{i}}  p(m_{i}|w_{1}, ..., w_{i-1}, w_{i+1}, ..., w_{n}).
\end{equation*}
$b_{i}$ is considered as literal, because according to relevant statistics from \citet{cameron2003metaphor}; \citet{martin2006corpus}; \citet{steen2010method} and \citet{shutova2016design}, literals are more common in typical corpora. Thus, a literal has higher probability appearing in a context than a metaphor with the similar meaning. Noticeably, the limitation of paraphrasing a metaphor with a single literal word is that the paraphrase may lose some nuance in the original metaphor. However, such a paraphrase method can help a machine to better translate the real meanings of metaphors into human comprehensible languages (see \S~\ref{sect: result}). 


\section{Dataset}
\label{sect: Dataset}

\noindent \textbf{MOH.}~~~MOH was formed by \citet{mohammad2016metaphor}, sourcing from WordNet example sentences. 
We select 315 sentences containing 315 metaphoric verbs whose metaphoricity was agreed by at least 70\% annotators, forming our automatic evaluation test set (MOH$_{315}$). The average sentence length is 8.8. We conduct human evaluation with 50 randomly selected sentences (MOH$_{50}$) from MOH$_{315}$.

\noindent \textbf{VUA$_{50}$.}~~~We also randomly select 50 metaphoric verbs and their associated sentences from VU Amsterdam Metaphor Corpus \citep{steen2010method} for human evaluation, where the sentences are originated from British National Corpus \citep{burnard2000reference}. 
VUA$_{50}$ may contain multiple metaphoric verbs in a sentence. Thus, 36 different sentences from all four different genres e.g., academic text, conversation, news and fiction are selected. The average sentence length of VUA$_{50}$ is 27.7.

\section{Baselines}
\label{sect: baseline}
\textbf{$\textsc{sim-cbow}_{I+O}$} \citep{mao2018word} modelled word co-occurrences with CBOW based word2vec \citep{mikolov2013distributed} input and output vectors. 
Their paraphrases were also constrained by WordNet hypernyms and synonyms.

\noindent \textbf{$\textsc{bert-large-cased-wwm}$} \citep{cui2019pre} masked whole words during the BERT pre-training procedure, rather than WordPieces. We test this method as an alternative of $\textsc{bert-large-cased}$ for the best fit word prediction. 
\begin{center}
\begin{figure*}[th!]
    \centering
    \includegraphics[scale=0.56]{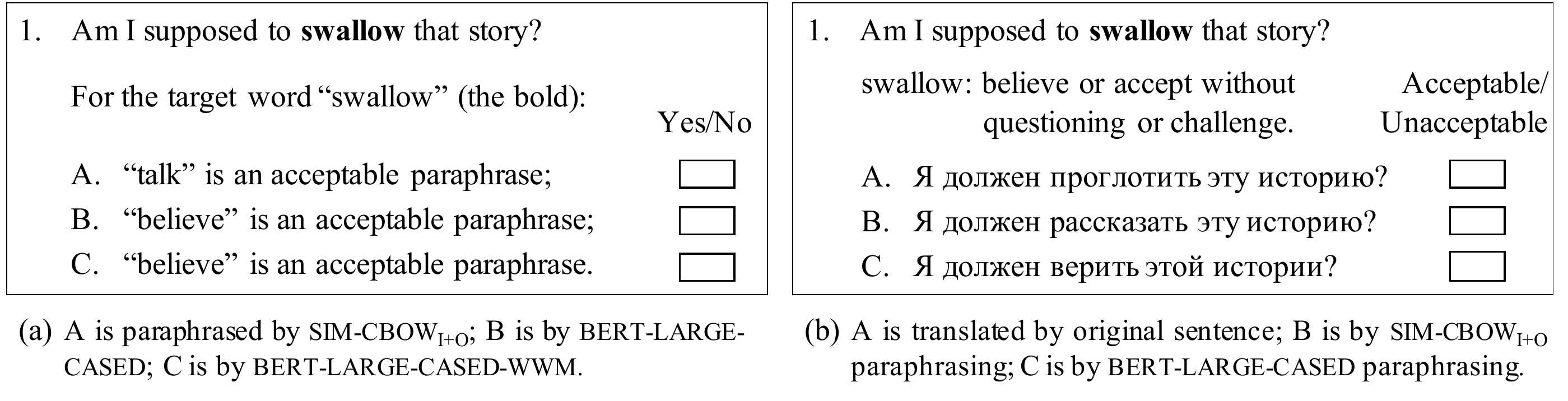}
    \caption{Example questions for human evaluation. (a) Paraphrasing task. (b) Machine Translation task.}
    \label{fg:questionnaire}
\end{figure*}
\end{center}
\vspace{-1.5em}
\section{Result} \label{sect: result}

The evaluation is conducted as three phases. First, we automatically evaluate model performance on MOH$_{315}$. Since MOH$_{315}$ originated from WordNet example sentences, a paraphrasing prediction is correct, if the paraphrase belongs to the WordNet sense class which the test sentence was in.

As seen under the first column in Table~\ref{tb: result}, the accuracy of $\textsc{bert-large-cased}$ on the automatic evaluation are 49\%, outperforming $\textsc{sim-cbow}_{I+O}$ by 8\%. It shows that using BERT can better paraphrase a target word in a sentence, based on constrained missing word prediction. This is likely because Transformer based BERT can better model the long-term dependency and word orders in sentences, while a shallow neural network based CBOW only represents word co-occurrences within a window of contexts. In CBOW, the position of a target word was not specified in a sentence, while BERT used positional embedding for each word during its pre-training procedure. Compared with $\textsc{bert-large-cased-wwm}$, there is a gain of 3\% in $\textsc{bert-large-cased}$. Masking whole words during pre-training does not outperform the original BERT that masked WordPieces,  probably because the learned stem information from WordPiece is more useful for predicting a missing word than the whole-word-masking approach. Thus, the original BERT can better retrieve a missing word.

\begin{table*}[ht!]
\begin{center} 
\begin{tabular}{c|cc|ccc} 
\Xhline{2.5\arrayrulewidth}
\multirow{2}{*}{\bf Model} & \multicolumn{2}{c|}{\bf Automatic Evaluation} & \multicolumn{3}{c}{\bf Human Evaluation}\\
 & \bf MOH$_{315}$ & \bf MOH$_{50}$ & \bf MOH$_{50}$ & \bf VUA$_{50}$ & \bf MOH$_{50}$+VUA$_{50}$\\ \hline
$\textsc{sim-cbow}_{I+O}$ & 0.41 & 0.42 & 0.68 & 0.70 & 0.69 \\
$\textsc{bert-large-cased-wwm}$ & 0.46 & 0.46 & 0.76 & 0.80 & 0.78 \\
$\textsc{bert-large-cased}$ & \bf 0.49 & \bf 0.50 & \bf 0.82 & \bf 0.82 & \bf 0.82\\

\Xhline{2.5\arrayrulewidth}
\end{tabular}
\end{center}
\caption{\label{tb: result} Metaphor interpretation evaluation on paraphrasing tasks, measured by accuracy. }
\end{table*}

\begin{table*}[t!]
\begin{center}
\begin{tabular}{>{\centering\arraybackslash}p{3.2cm}|>{\centering\arraybackslash}p{0.9cm}>{\centering\arraybackslash}p{0.9cm}>{\centering\arraybackslash}p{0.9cm}>{\centering\arraybackslash}p{0.9cm}>{\centering\arraybackslash}p{0.9cm}|>{\centering\arraybackslash}p{0.9cm}>{\centering\arraybackslash}p{0.9cm}>{\centering\arraybackslash}p{1.1cm}|>{\centering\arraybackslash}p{0.9cm}}
\Xhline{2.5\arrayrulewidth}
& German & Russian & Greek & Italian & Dutch & Chinese & Thai & Japanese & Avg\\
Cohen's kappa ($\kappa$) & 0.59 & 0.67 & 0.60 & 0.63 & 0.67 & 0.69 & 0.71 & 0.55 & - \\
\hline
Original & 0.48 & 0.32 & 0.32 & 0.50 & 0.50 & 0.38 & 0.30 & 0.28 & 0.385 \\
$\textsc{sim-cbow}_{I+O}$ & 0.38 & 0.40 & 0.36 & 0.56 & 0.46 & 0.60 & 0.52 & 0.44 & 0.465 \\
$\textsc{bert-large-cased}$ & \bf 0.60 & \bf 0.42 & \bf 0.46 & \bf 0.72 & \bf 0.66 & \bf 0.68 & \bf 0.70 & \bf 0.50 & \bf 0.594 \\
\Xhline{2.5\arrayrulewidth}
\end{tabular}
\end{center}
\caption{\label{tb: translation} Metaphor interpretation evaluation on Machine Translation tasks, measured by accuracy.}
\end{table*}

In the second phase, we test model performance on MOH$_{50}$ and VUA$_{50}$ with human evaluation. As seen in Table~\ref{tb: result}, the accuracy of automatic evaluation on MOH$_{50}$ given by $\textsc{sim-cbow}_{I+O}$, $\textsc{bert-large-cased-wwm}$ and $\textsc{bert-large-cased}$ is similar to their performance on MOH$_{315}$. For human evaluation, we invited 3 native English speakers with master degrees. A participant would be asked if a lemmatized paraphrase of a target word is acceptable in the given context with a questionnaire. A paraphrase is acceptable, if it captures the original meaning of a target word in the given context, where the grammatical tense of the lemmatized paraphrase is not a part of consideration.
An example of the questionnaire can be viewed in Figure~\ref{fg:questionnaire}a. 
The final decision is agreed by at least 2 annotators.

In the human evaluation (Table~\ref{tb: result}), $\textsc{bert-large-}$ $\textsc{cased}$ surpasses $\textsc{sim-cbow}_{I+O}$ by 14\%, 12\% and 13\% on MOH$_{50}$, VUA$_{50}$ and their combination datasets (Cohen's $\kappa=0.61$), respectively, yielding an average gain of 13\%. The average gain of $\textsc{bert-large-cased}$ over $\textsc{bert-}$ $\textsc{large-cased-wwm}$ is 4\%. The accuracy of the three models on MOH$_{50}$ with human evaluation is higher than their automatic evaluation, because candidate words from other WordNet sense classes are possibly acceptable from practical aspects. E.g., ``read'' is an acceptable paraphrase of ``scan'' in ``She \underline{scanned} the newspaper headlines while waiting for the taxi'' for human, although it is incorrect for automatic evaluation.  $\textsc{bert-large-cased}$ performs similarly on MOH$_{50}$ and VUA$_{50}$, although the average length of VUA sentences (27.7) is longer than that of MOH (8.8).

In the third phase, for evaluating the quality of paraphrases on a downstream task, we examine 
the MOH$_{50}$ paraphrases on English MT with 8 target languages, namely \textit{German}, \textit{Russian}, \textit{Greek}, \textit{Italian}, \textit{Dutch}, \textit{Chinese}, \textit{Thai} and \textit{Japanese}.
We invited 3 participants per target language, to annotate the quality of a translation of each target word in an original sentence, a $\textsc{sim-cbow}_{I+O}$ paraphrased sentence, and a $\textsc{bert-large-cased}$ paraphrased sentence. $\textsc{bert-large-cased-wwm}$ is excluded in this phase, because many predictions of it are the same as the predictions given by $\textsc{bert-large-cased}$ by our observation (e.g., B and C in Figure~\ref{fg:questionnaire}a), whereas $\textsc{bert-large-cased}$ surpasses $\textsc{bert-large-cased-wwm}$ on the paraphrase evaluations. The translations are given by Google translator. The participants are native in the target languages, and had learnt English in school, and had each spent a minimum of one year living and studying in England as a postgraduate. Participants were provided with senses of the original metaphors in WordNet, so that they can fully understand the true meanings of the metaphors. The quality of a translation is measured by a binary choice, e.g., acceptable or unacceptable. An acceptable translation is qualified, if a translation expresses the true meaning of the target word and it matches its context (cohesion) in their mother language, where errors in context words and tenses are ignored. The final decision was agreed by at least two annotators ($\kappa$ ranges from 0.55 to 0.71 in Table~\ref{tb: translation}).  
An example can be viewed in Figure~\ref{fg:questionnaire}b. 

As seen in Table~\ref{tb: translation}, the average accuracy of original metaphor translations is 38.5\%. By paraphrasing the metaphors with $\textsc{bert-large-cased}$, the average accuracy is improved to 59.4\%, achieving a gain of 20.9\%, outperforming $\textsc{sim-cbow}_{I+O}$ by 12.9\%. The average gain of $\textsc{bert-large-cased}$ on the target Asian languages (30.7\%) against the original metaphor translations are higher than that of European languages (15.0\%). This is probably because the larger language and cultural differences mean that English metaphors which are directly translated are unlikely to be understandable by Asians. After paraphrasing, the incomprehensible phrases become comprehensible. E.g., ``swallow'' in ``Am I supposed to \underline{swallow} that story?'' and ``tugged'' in ``She \underline{tugged} for years to make a decent living.'' are badly translated by Google across all the target languages. However, the translations based on $\textsc{bert-large-cased}$ paraphrased target words, e.g. ``Am I supposed to \underline{believe} that story?'' and ``She \underline{struggled} for years to make a decent living.'' are acceptable in each target language. 
We also noticed that there are some translation errors introduced due to incorrect paraphrasing. 
However, the accuracy gain achieved by paraphrasing metaphors substantially surpasses the small impact from the error, yielding a significant overall improvement (20.9\% in accuracy) in the MT task.


\section{Conclusion}
We propose an unsupervised model for metaphor interpretation based on BERT and WordNet, yielding large gains against the baseline on metaphor paraphrasing tasks.
Our evaluation demonstrates that paraphrasing English metaphors into their literal counterparts can help a MT system improve the accuracy of translating the metaphors into more comprehensible target languages. In future work, we will test our model on other downstream tasks. 

\bibliography{anthology,emnlp2020}
\bibliographystyle{acl_natbib}

\end{document}